\documentclass[journal]{IEEEtran}
\usepackage{graphicx,float,amssymb,multirow,pifont,amsmath,scalerel,makecell}
\usepackage{bm,hyperref,xcolor,enumitem,booktabs,arydshln,footnote,url,threeparttable}
\usepackage[caption=false]{subfig}
\usepackage{color}
\usepackage{cite}
\usepackage[numbers,sort&compress]{natbib}
\usepackage{amsthm,amsmath,amssymb}
\usepackage{mathrsfs}
\usepackage{threeparttable}

\usepackage[subfigure,titles]{tocloft}
\hypersetup{hypertex=true,
	colorlinks=true,
	linkcolor=blue,
	anchorcolor=blue,
	citecolor=blue}

\graphicspath{{Figures/}}

\begin{document}
\title{PPMamba: A Pyramid Pooling Local Auxiliary SSM-Based Model for Remote Sensing Image Semantic Segmentation}
\author{Yin Hu, 
    Xianping~Ma,
    Jialu~Sui,
	and Man-On~Pun,~\IEEEmembership{Senior Member,~IEEE,}
	\thanks{This work was supported in part by the Guangdong Provincial Key Laboratory of Future Networks of Intelligence under Grant 2022B1212010001 and National Natural Science Foundation of China under Grant 42371374 and 41801323. \textit{(Corresponding author: Man-On Pun)}}
	\thanks{Yin Hu, Xianping Ma, Jialu Sui and Man-On Pun are with the School of Science and Engineering, The Chinese University of Hong Kong, Shenzhen 518172, China (e-mail: yinhu@link.cuhk.edu.cn; xianpingma@link.cuhk.edu.cn; jialusui@link.cuhk.edu.cn; SimonPun@cuhk.edu.cn).}}
\maketitle

\begin{abstract}
Semantic segmentation is a vital task in the field of remote sensing (RS). However, conventional convolutional neural network (CNN) and transformer-based models face limitations in capturing long-range dependencies or are often computationally intensive. Recently, an advanced state space model (SSM), namely Mamba, was introduced, offering linear computational complexity while effectively establishing long-distance dependencies. Despite their advantages, Mamba-based methods encounter challenges in preserving local semantic information. To cope with these challenges, this paper proposes a novel network called Pyramid Pooling Mamba (PPMamba), which integrates CNN and Mamba for RS semantic segmentation tasks. The core structure of PPMamba, the Pyramid Pooling-State Space Model (PP-SSM) block, combines a local auxiliary mechanism with an omnidirectional state space model (OSS) that selectively scans feature maps from eight directions, capturing comprehensive feature information. Additionally, the auxiliary mechanism includes pyramid-shaped convolutional branches designed to extract features at multiple scales. Extensive experiments on two widely-used datasets, ISPRS Vaihingen and LoveDA Urban, demonstrate that PPMamba achieves competitive performance compared to state-of-the-art models.
\end{abstract}


\IEEEpeerreviewmaketitle

\section{Introduction}\label{sec:int}
The rapid development of remote sensing (RS) technologies has dramatically transformed our understanding of time and space scales on the Earth. RS technologies are extensively applied in agriculture \citep{wojtowicz2016application}, forestry \citep{lechner2020applications}, geology \citep{bhan1983applications}, meteorology \citep{yates1975meteorological}, military \citep{hudson1975military} and environmental protection \citep{zhao2017overview}, enabling systematic analyses, assessments, and predictions. Among these applications, semantic segmentation, which assigns class labels to each pixel in an image, serves as a foundation for many downstream geoscientific tasks, such as land cover classification and urban expansion monitoring  \citep{diakogiannis2020resunet, ma2023sam}. 

In recent years, deep learning has significantly advanced the performance of semantic segmentation in RS, primarily due to its ability to extract abstract and hierarchically structured features from RS images \citep{zhu2017deep}. Convolutional neural network (CNN) and transformer are the most commonly used techniques in state-of-the-art deep learning models. CNN-based models \citep{he2016deep, li2021abcnet, wu2023cmtfnet} excel at capturing local information through convolution operations, while transformer-based models \citep{ma2023unsupervised, chen2021transunet, dosovitskiy2021an} leverage self-attention mechanisms \citep{vaswani2017attention} to model long-distance dependencies. However, these methods still have limitations in RS applications. CNN-based models struggle to capture global context due to their restricted receptive fields, while transformers, although capable of modeling long-range dependencies, face significant computational challenges when handling high-resolution, large-scale RS data \citep{dosovitskiy2021an}.

To overcome these challenges, Mamba, a novel state space model (SSM)-based network, was introduced \citep{gu2023mamba}, offering a promising solution to effectively capture long-distance dependencies with linear computational complexity. Various SSM-based models have been successfully applied across different domains, including Vmamba \citep{liu2024vmamba} and Vision Mamba \citep{zhu2024vision} in computer vision, as well as RSMamba \citep{chen2024rsmamba} and RS3Mamba \citep{ma2024rs3mamba} in RS. Innovations such as Mamba-in-Mamba \citep{zhou2024mamba} for hyperspectral image classification, Pan-Mamba \citep{he2024pan}, and ChangeMamba \citep{chen2024changemamba} for RS pan-sharpening and change detection have also emerged. Despite the advantages of these models, they struggle to characterize local details, which is critical for accurate RS image segmentation.

This paper proposes Pyramid Pooling Mamba (PPMamba), a novel network designed to address the local information loss in existing SSM-based models for RS image semantic segmentation. PPMamba consists of several layers of Pyramid Pooling-State Space Model (PP-SSM) blocks, and each block constructs multi-branch convolution-based blocks to assist the model in capturing features from each image patch. Additionally, the auxiliary multi-branch convolution-based blocks are structured in a pyramid shape in order to capture features at different scales. Since the land cover patterns in RS images are oriented in various spatial directions, the model possesses an omnidirectional state space (OSS) block to maximally establish long-distance dependencies. The structure of PP-SSM consists solely of Mamba and convolution-based blocks, leading to the capability of learning long-range dependencies with linear computational complexity. Extensive experiments on two widely used datasets, ISPRS Vaihingen and LoveDA Urban \citep{junjue_wang_2021_5706578, loveda}, validate the effectiveness of PPMamba. The results show that PPMamba outperforms several state-of-the-art models, highlighting its potential to address the unique challenges of RS image semantic segmentation. The main contributions of this article can be summarized as follows: 

\begin{enumerate}
\item A novel Mamba-based network, PPMamba, is proposed to effectively model local and global relationships in RS images while maintaining linear computational complexity. By integrating CNN-based pyramid pooling and the Mamba model, PPMamba addresses the limitations of existing methods in balancing fine-grained local feature extraction with comprehensive global context modeling.

\item The core structure of PPMamba, the PP-SSM block, introduces a pyramid-shaped convolutional module combined with OSS. This block effectively fuses multi-scale local features, selectively scanned from eight different directions, with global features, enhancing the model's ability to capture diverse land cover patterns in RS images.
\end{enumerate}

The remainder of this paper is organized as follows. Section~\ref{sec:relatedwork} reviews the related works on architectures and techniques relevant to PPMamba, while Section~\ref{sec:methodology} details the proposed method. Section~\ref{sec:experiments} presents the experimental results and discussions, followed by the conclusion in Section~\ref{sec:conclusion}.

\underline{Notation}: Vectors and matrices are denoted by bold-face letters. ${\bm I}_N$ is the $N \times N$ identity matrix while $\left[\cdot\right]^T$ denotes the transposition of the enclosed vector.

\section{Related Work }\label{sec:relatedwork}
\subsection{Remote Sensing Image Semantic Segmentation} Early approaches for RS image semantic segmentation primarily relied on traditional image processing techniques and classical machine learning algorithms. Methods such as pixel-level classification were widely adopted, with techniques like the Maximum Likelihood Classifier (MLC) \citep{strahler1980use} and Support Vector Machine (SVM) \citep{mountrakis2011support} being popular due to their simplicity and effectiveness. However, these methods typically struggled to capture spatial information and often underperformed when dealing with complex object categories in high-dimensional data.

With the emergence of deep learning, CNN and transformer-based models have demonstrated significant potential in RS image segmentation \citep{ma2022crossmodal, ma2024multilevel, sui2023gcrdn}. CNN-based models, such as ResUNet-a \citep{diakogiannis2020resunet}, leverage hierarchical feature extraction through convolutional layers and have been enhanced with techniques like residual connections and pyramid scene parsing. However, CNNs are limited by their local receptive fields, making it challenging to capture long-range dependencies. To address this problem, transformer-based models, such as GLOTS \citep{liu2023rethinking}, have been introduced, utilizing self-attention mechanisms to capture global context. Despite their strengths, transformers are computationally intensive, leading to high resource demands for processing high-resolution RS images \citep{sui2024diffusion}. These challenges highlight the need for new architectures that balance segmentation accuracy and computational efficiency.

\begin{figure*}[t]
	\centering
	\includegraphics[width=\linewidth]{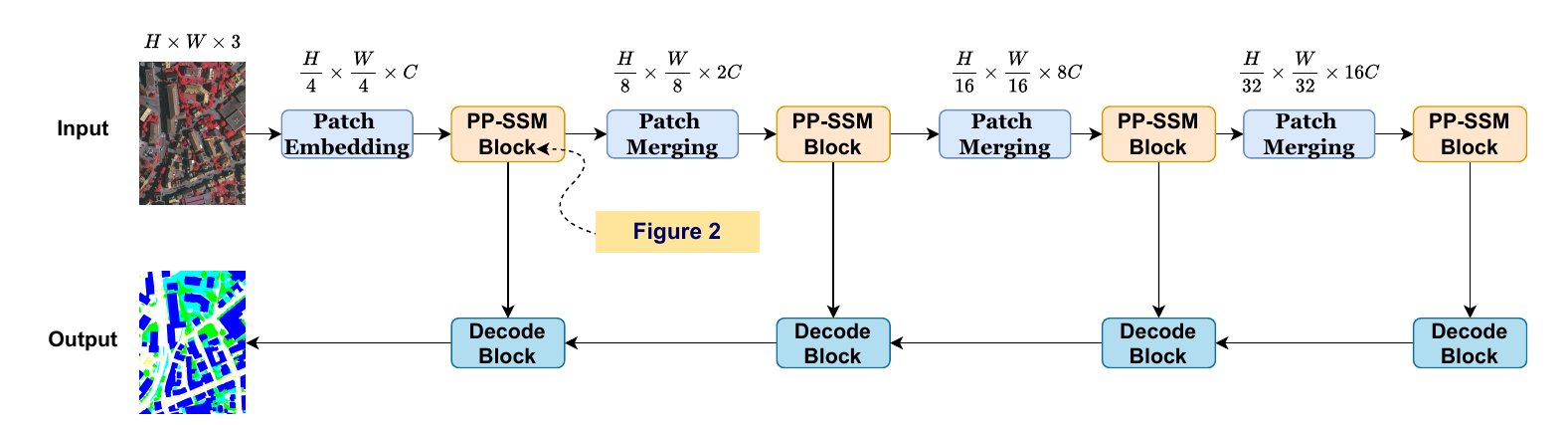}
	\caption{The architecture of proposed PPMamba.}
	\label{fig1}
\end{figure*}

\subsection{Mamba} 
The Mamba architecture was introduced as an alternative to transformers, addressing their high computational complexity while capturing long-range dependencies in visual data. Mamba is based on the structured state space model (SSM), originally designed to handle continuous data with linear time complexity \citep{gu2021efficiently}. The transition from SSM to structured state space sequence models (S4) allowed for effective processing of discrete data \citep{gu2021efficiently}. More specifically, we consider a continuous system that maps a 1-D function or sequence $x(t)\in \mathbb{R}\mapsto y(t) \in \mathbb{R}$ through a hidden state ${\bm h}(t)\in \mathbb{R} ^{N\times 1}$. This process can be described as a linear Ordinary Differential Equation (ODE) \citep{gu2021efficiently}: 
\begin{equation}
	\begin{split}
		{\bm h}'(t)&={\bm A}{\bm h}(t)+{\bm b} x(t),\\
		y(t)&={\bm c}^T{\bm h}(t),
	\end{split}
\end{equation}
where ${\bm A}\in\mathbb{R}^{N\times N}$ denotes the state transition matrix while ${\bm b}\in\mathbb{R}^{N\times 1}$ and ${\bm c}\in\mathbb{R}^{N\times 1}$ are the projection parameters. Furthermore, ${\bm h}'(t)$ stands for the derivative of ${\bm h}(t)$. To adapt the system to a discrete form, a zero-order hold (ZOH) is required to convert all the parameters into their discrete counterparts, as follows:
\begin{equation}
	\begin{split}
		\bar{\bm A}&=\exp(\bigtriangleup \bm A),\\
		\bar{\bm b}&=(\bigtriangleup \bm A)^{-1} (\exp(\bigtriangleup \bm A)- {\bm I}_N)\cdot\bigtriangleup {\bm b},
	\end{split}
\end{equation}
where $\bigtriangleup$ is a step size that denotes the input's resolution, $\bar{\bm A}$ and $\bar{\bm b}$ are the discrete version of the projection parameters $\bm A$ and $\bm b$, respectively. 

However, the S4 model faced challenges in optimizing computational efficiency, which led to the development of the selective structured state space model (S6) \citep{gu2023mamba}. S6 forms the core of Mamba, introducing dynamic adjustments to $\bm b$, $\bm c$, and $\bigtriangleup$ that depend on the input, enabling hardware-aware optimizations and selective compression of information.

Recently, numerous SSM-based models have been applied across various domains, including computer vision and remote sensing. In computer vision, Vmamba and Vision Mamba have introduced innovative approaches leveraging SSM-based architectures. Vmamba maintains linear complexity while preserving global receptive fields by incorporating a Cross-Scan Module (CSM) that traverses the spatial domain and transforms non-causal visual images into ordered patch sequences \citep{liu2024vmamba}. Furthermore, Vision Mamba demonstrates that self-attention is not necessary for visual learning by exploiting bidirectional Mamba blocks with position embeddings to structure images and bidirectional state space models for compression \citep{zhu2024vision}. In remote sensing, RSMamba presents an innovative architecture for image classification, introducing a dynamic multi-path activation mechanism to enhance Mamba’s capability in modeling non-causal data \citep{chen2024rsmamba}. Recently, Pan-Mamba was developed to perform cross-modal information exchange by integrating channel swapping and cross-modal Mamba designs, enabling efficient fusion across modalities \citep{he2024pan}. Additionally, Mamba-in-Mamba has shown strong performance in hyperspectral image classification \citep{zhou2024mamba}, while ChangeMamba pioneers the application of the Mamba architecture for RS change detection tasks \citep{chen2024changemamba}. Despite these advancements, most of the above models are not explicitly designed for semantic segmentation. To address this challenge, RS3Mamba was proposed as one of the earliest SSM-based models tailored for RS image semantic segmentation \citep{ma2024rs3mamba}. Following this, PyramidMamba introduced an adaptable decoder featuring dense spatial pyramid pooling (DSPP) to capture multiscale semantic features \citep{wang2024pyramidmamba}. However, RS3Mamba’s intricate architecture imposes significant computational overhead, and PyramidMamba’s emphasis on pyramid pooling in the decoder may result in suboptimal multiscale feature extraction within its encoder.

\begin{figure*}[t]
	\centering
	\includegraphics[width=0.8\linewidth]{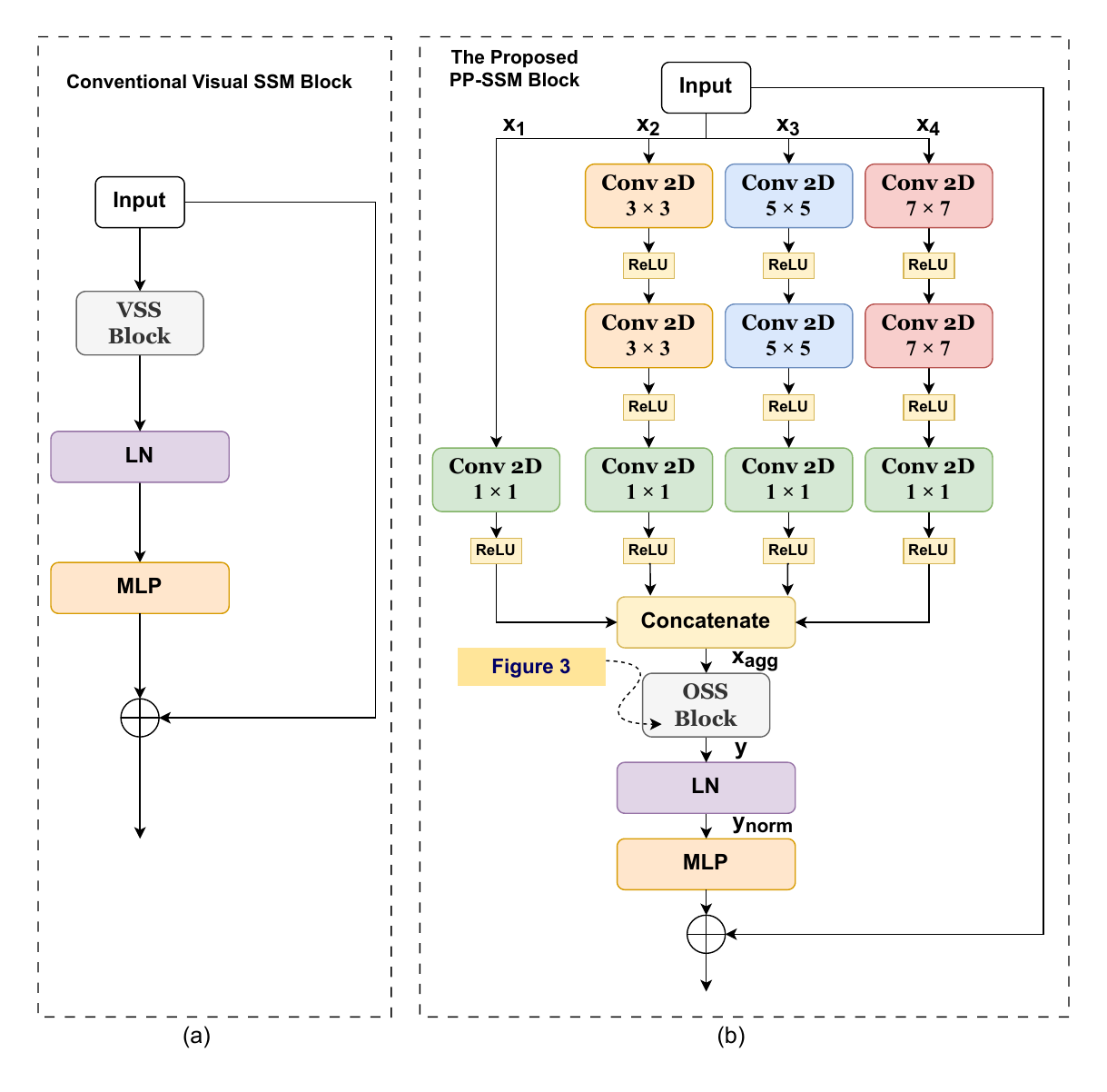}
	\caption{The architectures of the conventional visual SSM block and the proposed PP-SSM block. (a) The architecture of a conventional visual SSM block. (b) Proposed architecture of PP-SSM Block.}
	\label{fig2}
\end{figure*}

\subsection{Spatial Pyramid Pooling}
Spatial Pyramid Pooling (SPP) was developed to address the rigid input size requirements of early CNN architectures, enabling models to handle variable input sizes without losing critical spatial information \citep{he2015spatial}. By introducing multilevel pooling operations, SPP allows models such as AlexNet \citep{krizhevsky2012imagenet} and VGGNet \citep{simonyan2014very} to preserve spatial hierarchies while generating fixed-length output vectors. This capability has proven essential in high-resolution image tasks, where resizing can distort important features. In RS image segmentation, SPP has been widely adopted for multiscale feature extraction, providing flexibility in adapting to the diverse spatial patterns found in RS imagery. Advanced architectures, such as Faster R-CNN \citep{girshick2015fast} and YOLO \citep{redmon2016you}, have integrated SPP to enhance their segmentation accuracy by better capturing context across different scales. Despite these advances, current models often emphasize either local detail (as in CNN-based approaches) or global context (as in transformer-based models), leading to suboptimal performance in scenarios requiring a nuanced understanding of both. The challenge remains to develop an architecture that effectively integrates multiscale local and global features while maintaining computational efficiency.

\section{Methodology }\label{sec:methodology}
\subsection{Proposed PPMamba} 
The proposed PPMamba architecture is illustrated in Fig.~\ref{fig1}. The input to the model is an image with dimensions $H\times W \times 3$, processed through a UNet-like encoder-decoder framework. The encoder reduces the spatial resolution of the input while preserving essential features. Furthermore, the decoder progressively upsamples the features to produce the final segmentation map. In the encoder, the input image first undergoes a patch embedding operation, converting it into feature maps of size $\frac{H}{4}\times\frac{W}{4}\times C$. These feature maps are then passed through a sequence of patch merging operations and PP-SSM blocks. The patch merging operations successively reduce the spatial resolution from $\frac{H}{4}\times\frac{W}{4}$ to $\frac{H}{32}\times\frac{W}{32}$, while increasing the number of channels to $16C$. The stacked PP-SSM blocks enable the model to capture both local and global context information while maintaining computational efficiency. The decoder consists of four stages of upsampling. Each decode block fuses the upsampled features with the corresponding encoder features and features from its previous decode block, enabling the reconstruction of detailed spatial information. The output is a high-resolution segmentation map with dimensions $H \times W \times 3$.

\begin{figure*}[t]
	\centering
	\includegraphics[width=\linewidth]{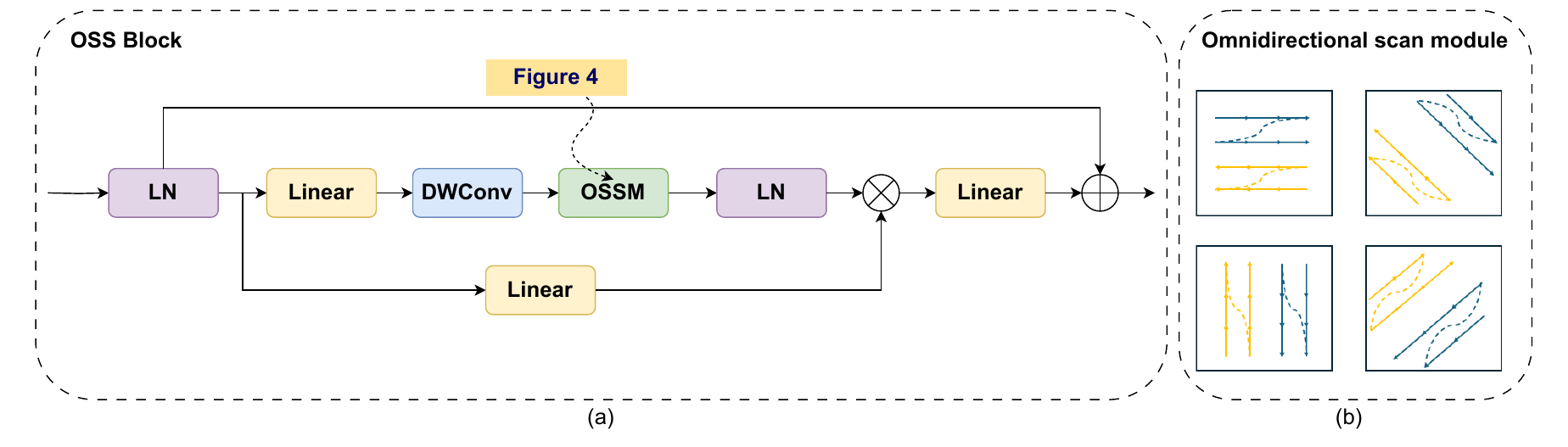}
	\caption{(a) The architecture of the proposed OSS block. (b) The illustration of the selective scan directions of OSSM.}
	\label{fig3}
\end{figure*}

\subsection{Proposed PP-SSM Block}

Fig.~\ref{fig2}(a) shows the structure of a conventional visual SSM block in which input is processed by visual state space (VSS) blocks followed by a layer normalization (LN) block and a multilayer perceptron (MLP) block. However, the VSS block suffers from many limitations in capturing global spatial features from RS images. 

In sharp contrast, the proposed PP-SSM block, shown in Fig.~\ref{fig2}(b), is the core structure in our PPMamba model, utilizing a multi-branch auxiliary methodology for RS image semantic segmentation. First, the input is separated into four distinct parts along the channel dimension, namely ${\bm x}_{1} , {\bm x}_{2} , {\bm x}_{3}$ and ${\bm x}_{4}$, as shown in  Fig.~\ref{fig2}(b). This separation allows the PP-SSM block to independently capture different aspects of the local features using four SPP branches. These SPP branches stack continuous convolutional layers with different kernel sizes to capture the local features while maintaining the input's resolution the same way as the output to preserve the local spatial information. Specifically, ${\bm x}_{2} , {\bm x}_{3}$ and ${\bm x}_{4}$ are passed through two layers of convolutional blocks of kernel sizes  $3\times 3$, $5\times 5$ and $7\times 7$ respectively, and the ReLU activation functions. Finally, the resulting features together with ${\bm x}_{1}$ are processed by 2D convolutional blocks of kernel size $1\times 1$.

It is worth pointing out that employing various kernel sizes to process ${\bm x}_{2} , {\bm x}_{3}$ and ${\bm x}_{4}$ can form a pyramid structure, enabling the model to capture a wider range of local features at different scales. The pyramid-shaped design is crucial for extracting comprehensive local features from the input image, which is essential for accurate semantic segmentation. The output of each convolution-based block will be passed through a ReLU activation that introduces non-linearity into the model and enhances its capability to learn complex patterns from the input data. 

After processing through the convolutional layers, the PP-SSM block concatenates the outputs to form a unified feature map ${\bm x}_{agg}$ with the same number of channels as the original input. After that, ${\bm x}_{agg}$  is input to an omnidirectional state space block (OSS) \citep{zhao2024rs} to capture the global features of the RS images. The OSS block performs selective scanning in multiple directions to capture global dependencies and spatial relationships from various angles. Detailed operations of the OSS will be elaborated in the next section. The output of OSS, denoted as ${\bm y}$, is first normalized before being processed by an MLP block. The normalization block makes the training process converge faster, while the MLP block can adjust the input dimensions.

In summary, the PP-SSM block introduces four convolution-based branches with various kernel sizes to collect local features. Furthermore, the pyramid-shaped kernel sizes capture features across different dimensions.

\begin{figure*}[t]
	\centering
	\includegraphics[width=0.9\linewidth]{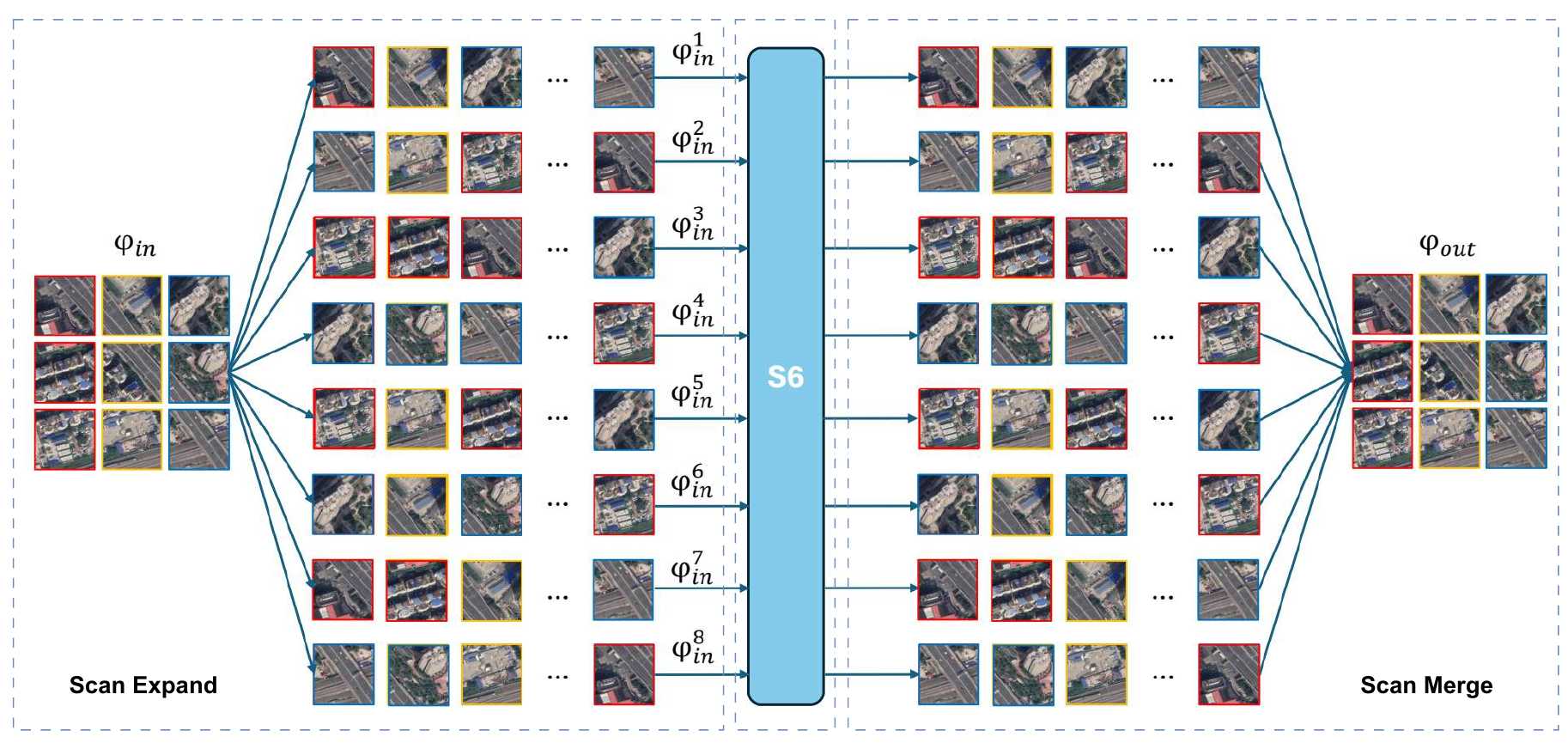}
	\caption{Illustration of the operation of the proposed oriented scanning module (OSSM).}  
	\label{fig4}
\end{figure*}

\subsection{Omnidirectional State Space Block (OSS)}

\begin{table*}[h]
	\begin{center}
		\caption{The performance of PPMamba and other state-of-the-art models on the Vaihingen dataset, where type C indicates CNN-based models, T indicates transformer-based models, C-T indicates CNN\&transformer-based models, and M indicates Mamba-based models. F1-score and IoU are chosen as evaluation metrics. The accuracy of each category is presented by F1/IoU. \textbf{Bold} font represents the best values.}
		\label{tab:table1}
		\begin{tabular}{ccccccccc}   
            \toprule
			\textbf{Model}&\textbf{Type}&\textbf{impervious surface}&\textbf{building}&
            \textbf{low vegetation}&\textbf{tree}&\textbf{car}&\textbf{mF1}&\textbf{mIoU} \\ 
            \midrule
			ABCNet \citep{li2021abcnet}& C & 89.68/90.45 & 93.72/93.90 & 77.93/75.52 & 89.81/91.07 & 73.46/63.16 & 84.92 & 74.57\\ 
			MANet \citep{li2021multiattention}& C & 90.28/91.74 & 94.28/93.07 & 78.95/\textbf{79.26} & 89.85/89.76 & 77.58/70.76 & 86.19 & 76.32\\ 
			CMTFNet \citep{wu2023cmtfnet}& C & 90.69/90.50 & 95.03/96.20 & 78.89/76.18 & 90.13/91.33 & 82.09/74.95 & 87.37 & 78.06\\ 
            FTUNetFormer \citep{wang2022unetformer}& T & 90.78/90.37 & 94.54/94.88 & 76.48/73.59 & 89.15/91.83 & 75.28/66.49 & 85.25 & 75.09 \\ 
            UNetFormer \citep{wang2022unetformer}& C-T & 90.37/\textbf{92.19} & 94.58/93.44 & 78.37/76.56 & 90.19/91.15 & 81.85/75.87 & 87.07 & 77.60 \\ 
            HST\_UNet \citep{zhou2024hybrid}& C-T & 91.27/91.34 & 95.36/95.43 & 78.44/77.27 & 90.04/91.02 & 83.61/79.07 & 86.62 & 78.67 \\ 
            TransUNet \citep{chen2021transunet}& C-T & 91.24/90.31 & 94.82/\textbf{96.63} & 78.85/74.71 & 90.54/92.79 & 83.77/78.97 & 87.84 & 78.78 \\
            RS3Mamba \citep{ma2024rs3mamba}& M & 90.87/89.99 & 95.26/95.59 & 78.49/75.74 & 90.20/91.93 & 81.83/74.10 & 87.33 & 78.04 \\
            RS-Mamba \citep{zhao2024rs}& M & 88.37/87.73 & 92.52/92.08 & 76.31/75.68 & 89.14/90.14 & 72.20/64.24 & 83.71 & 72.77 \\
            \midrule
            PPMamba& M & \textbf{91.86}/91.01 & \textbf{95.94}/96.52 & \textbf{79.04}/77.17 & \textbf{90.23}/\textbf{92.08} & \textbf{84.61}/\textbf{80.03} & \textbf{88.34} & \textbf{79.60} \\
            \bottomrule
		\end{tabular}
	\end{center}
\end{table*}

\begin{figure*}[t]
	\centering
	\includegraphics[width=\linewidth]{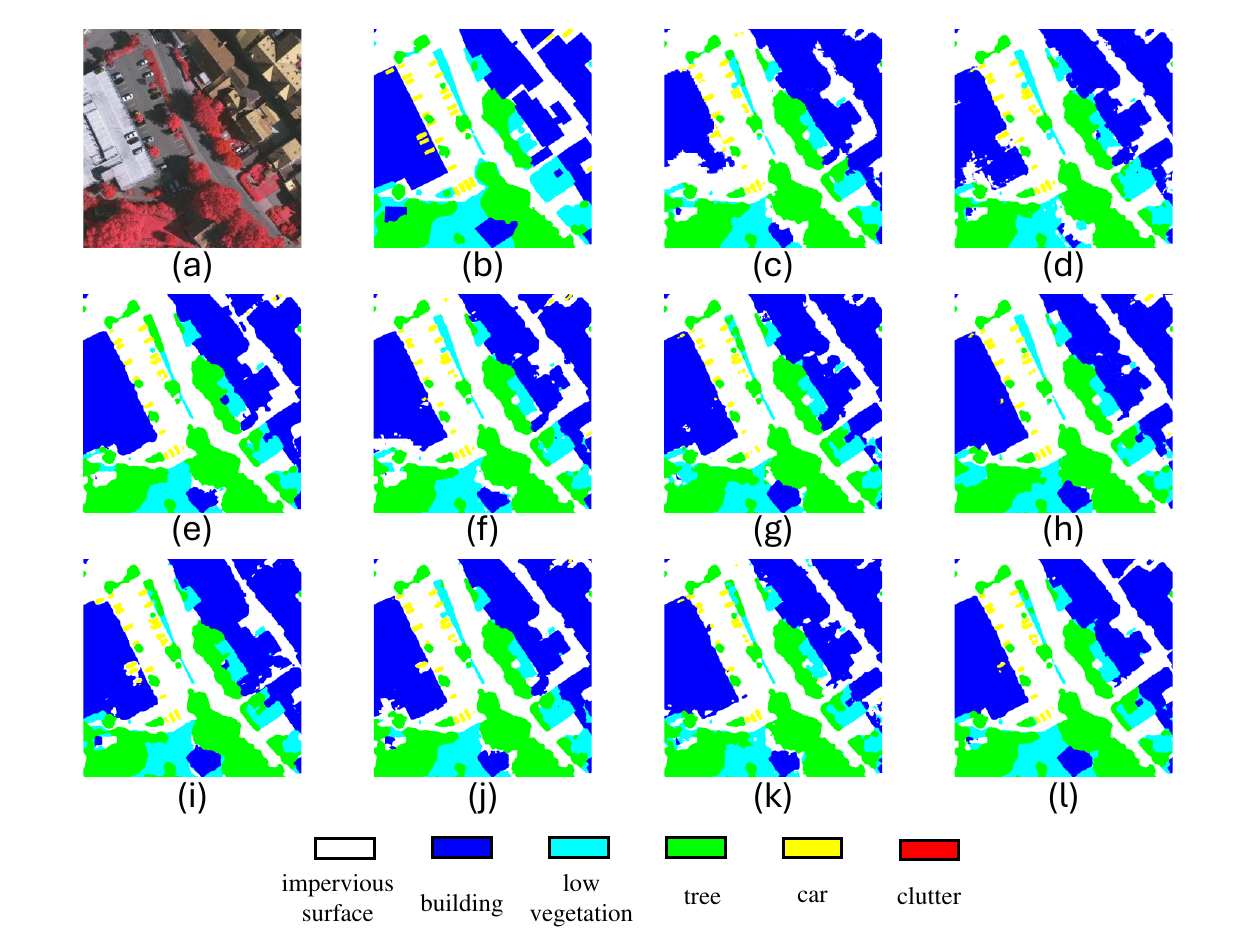}
	\caption{Performance comparisons on the ISPRS Vaihaigen dataset with the size of $1024 \times 1024$. (a) NIRRG images, (b) Ground truth, (c) ABCNet, (d) MANet, (e) CMTFNet, (f) UNetFormer, (g) FTUNetFormer, (h) HST\_UNet, (i) TransUNet, (j) RS3Mamba, (k) RS-Mamba and (l) PPMamba. }
	\label{fig5}
\end{figure*}

As shown in Fig.~\ref{fig3}(a), the architecture of the proposed OSS block begins with a layer normalization stage to stabilize the training process. Next, a linear transformation adjusts the input dimensions before the data passes through a depthwise convolution operation (DWConv) to extract spatial features. The core structure of the OSS block called the omnidirectional selective scan module (OSSM), then selectively scans the features in forward and backward directions across four different angles, i.e., eight scanning directions, as depicted in Fig.~\ref{fig3}(b). Finally, the output is passed through a linear transformation before the residual connections are applied to concatenate the input features with the final output.

The operation of OSSM is illustrated in Fig.~\ref{fig4}. We denote by $\varphi_{in}$ and $\varphi_{out}$ the input and output features of OSSM, respectively. The scanning process can be described as follows:
\begin{equation}
	\begin{split}
	\varphi_{in}^n &= expand(\varphi_{in},n),\\
	\varphi_{in}^n &= S6(\varphi_{in}^n), \\
    \varphi_{out} &= merge(\varphi_{in}^1,\varphi_{in}^2,\varphi_{in}^3,\varphi_{in}^4,\varphi_{in}^5,\varphi_{in}^6,\varphi_{in}^7,\varphi_{in}^8),
	\end{split}
\end{equation}
where $n \in N = \left \{ 1,2,3,...,8 \right \} $ represents the eight different scanning directions. Furthermore, $expand(\cdot)$ and $merge(\cdot)$ denote the scan expansion and merging operation, respectively. Finally, $S6(\cdot)$ is the selective scan space state sequential model \citep{gu2023mamba}. 

\section{Experiments}\label{sec:experiments}
\subsection{Datasets}\label{sec:dataset}
\subsubsection{ISPRS Vaihingen} 
The Vaihingen dataset consists of high-resolution aerial images captured over Vaihingen, Germany, as part of the German Association of Photogrammetry and Remote Sensing (DGPF) benchmark. The dataset contains 16 true orthophotos, each of resolution $2500\times 2000$ pixels. For our experiments, $12$ orthophotos were used as the training set, and the remaining 4 orthophotos were used for testing. The training set includes images with indices $1, 3, 23, 26, 7, 11, 13, 28, 17, 32, 34$, and $37$, while the test set comprises images with indices $5, 21, 15$, and $30$. Each orthophoto contains three spectral bands: near-infrared (NIR), red, and green (NIRRG). The ground sampling distance is $9$~centimeters, and the dataset is annotated with five foreground classes: impervious surfaces, buildings, low vegetation, trees, and cars, along with a background class.

\subsubsection{LoveDA Urban}\label{sec:lovedaurban}
The LoveDA dataset \citep{junjue_wang_2021_5706578, loveda} provides high-resolution RS images, with $5987$ samples in total, captured over three cities in China: Nanjing, Changzhou, and Wuhan. For this study, we focus on the urban subset, which includes 1833 images, each with a resolution of $1024\times 1024$ pixels. The dataset is split into 1156 training images and $677$ testing images. The training set consists of images indexed from No. 1366 to No. 2521, while the test set covers indices from No. 3514 to No. 4190. The images are provided in three channels: red, green, and blue (RGB), with a ground sampling distance of $30$~centimeters. The LoveDA Urban dataset includes seven land cover classes: background, buildings, roads, water, barren land, forests, and agriculture.

\subsection{Evaluation Metrics}\label{sec:evaluationmetrics}
Mean intersection over union (mIoU) and mean F1-score (mF1) were used to evaluate the performance of the models. Besides, $precision$ and $recall$ were used to calculate the F1-score. The definitions and equations for these metrics are as follows:
\begin{equation}
	\begin{split}
	Precision=\frac{TP}{TP+FP}, 
	\end{split}
\end{equation}
\begin{equation}
	\begin{split}
	Recall=\frac{TP}{TP+FN}, 
	\end{split}
\end{equation}
\begin{equation}
	\begin{split}
	F1-score = \frac{2(Precision\cdot Recall)}{Precision+Recall},\\
	mF1 = \frac{1}{k+1} \sum_{i=0}^{k} \frac{2(Precision\cdot Recall)}{Precision+Recall},
	\end{split}
\end{equation}
\begin{equation}
	\begin{split}
	IoU = \frac{TP}{FN+FP+TP},\\
	mIoU = \frac{1}{k+1} \sum_{i=0}^{k} \frac{TP}{FN+FP+TP},
	\end{split}
\end{equation}
where $k$ is the number of categories, $TP$ denotes true positives, $FP$ denotes false positives, and $FN$ denotes false negatives.

\begin{table*}[h]
	\begin{center}
		\caption{The performance of PPMamba and other state-of-the-art models on the LoveDA Urban dataset, where type C indicates CNN-based models, T indicates transformer-based models, C-T indicates CNN\&transformer-based models, and M indicates Mamba-based models. F1-score and IoU are chosen as evaluation metrics. The accuracy of each category is presented by F1/IoU. \textbf{Bold} font represents the best values.}
		\label{tab:table2}
		\begin{tabular}{ccccccccccc}   
			\toprule
			\textbf{Model}&\textbf{Type}&\textbf{background}&\textbf{building}&
			\textbf{road}&\textbf{water}&\textbf{barren}&\textbf{forest}&\textbf{agriculture}&\textbf{mF1}&\textbf{mIoU} \\ 
			\midrule
			ABCNet \citep{li2021abcnet}& C & 51.79/66.54 & 67.87/70.66 & 64.94/56.57 & 67.56/58.12 & 45.94/35.23 & 54.02/77.60 & \textbf{28.93}/\textbf{19.09} & 59.62 & 43.05\\ 
			MANet \citep{li2021multiattention}& C & 51.42/63.92 & 70.55/74.98 & 65.37/\textbf{64.73} & 70.17/64.60 & \textbf{48.33}/\textbf{40.67} & 52.69/84.92 & 5.22/2.81 & 61.17 & 44.72\\
			CMTFNet \citep{wu2023cmtfnet}& C & \textbf{52.57}/67.64 & 70.05/74.06 & 68.81/60.70 & 69.09/58.19 & 37.70/25.77 & 54.29/83.01 & 26.18/16.59 & 59.64 & 43.60\\
			FTUNetFormer \citep{wang2022unetformer}& T & 49.84/61.82 & 69.91/70.42 & \textbf{68.88}/64.03 & 67.73/56.98 & 26.81/16.82 & 51.40/89.96 & 23.29/14.18 & 56.63 & 41.23\\
			UNetFormer \citep{wang2022unetformer}& C-T & 51.57/64.48 & 69.10/70.94 & 64.40/61.48 & 67.07/65.24 & 44.49/34.20 & 54.23/82.87 & 16.99/9.95 & 59.32 & 42.82\\
			HST\_UNet \citep{zhou2024hybrid}& C-T & 50.39/68.61 & 71.07/70.02 & 70.57/61.95 & 68.67/58.01 & 17.67/10.01 & 54.00/84.64 & 27.76/16.70 & 55.68 & 41.06\\
			TransUNet \citep{chen2021transunet}& C-T & 52.47/67.23 & 67.19/60.93 & 67.03/59.46 & 73.68/63.13 & 40.59/31.84 & 43.75/85.72 & 0.00/0.00 & 60.19 & 44.07\\
			RS3Mamba \citep{ma2024rs3mamba}& M & 51.03/67.73 & 69.98/70.53 & 68.86/63.39 & 70.51/61.57 & 41.52/28.46 & \textbf{58.20}/85.67 & 23.59/14.50 & 60.38 & 44.25\\
			RS-Mamba \citep{zhao2024rs}& M & 48.83/64.63 & 60.62/55.34 & 59.64/53.95 & 67.92/54.16 & 35.36/26.26 & 47.85/\textbf{90.74} & 3.51/1.80 & 54.47 & 38.24\\
			\midrule
			PPMamba& M & 49.67/\textbf{69.27} & \textbf{71.25}/\textbf{77.44} & 68.64/60.28 & \textbf{78.35}/\textbf{66.87} & 40.88/28.35 & 53.43/79.67 & 6.86/3.57 & \textbf{61.76} & \textbf{46.14}\\
			\bottomrule
		\end{tabular}
	\end{center}
\end{table*}

\subsection{Implementation Details}\label{sec:implementationdetails}
Stochastic gradient descent (SGD) was applied as the optimization algorithm for training all models. The learning rate, momentum, and decaying coefficient values were set to $0.01, 0.9$, and $0.0005$, respectively. The batch size was set to $10$, while the epoch size $50$. The number of PP-SSM blocks at each stage is $[2, 2, 9, 2]$. No pre-trained strategy is loaded in order to confirm the effectiveness of the PPMamba architecture. Evaluation metrics were calculated twice per epoch. The experiments were conducted on a server node running Ubuntu 22.04.1 operating system, equipped with an NVIDIA GeForce RTX 4090 GPU. The framework utilized in these experiments was PyTorch 2.2.2.

\subsection{Performance Comparison}\label{sec:performancecomparison}
To evaluate the effectiveness of PPMamba, we conducted comparative experiments against nine state-of-the-art models. The baseline model used in these experiments is RS-Mamba \citep{zhao2024rs}. The comparison models include CNN-based methods, ABCNet \citep{li2021abcnet}, MANet \citep{li2021multiattention}, and CMTFNet \citep{wu2023cmtfnet}, transformer-based methods, FTUNetFormer \citep{wang2022unetformer}, hybrid CNN-transformer models, UNetFormer \citep{wang2022unetformer}, HST\_UNet \citep{zhou2024hybrid}, and TransUNet \citep{chen2021transunet}, and other Mamba-based models, RS3Mamba \citep{ma2024rs3mamba}.

\begin{figure*}[t]
    \centering
    \includegraphics[width=\linewidth]{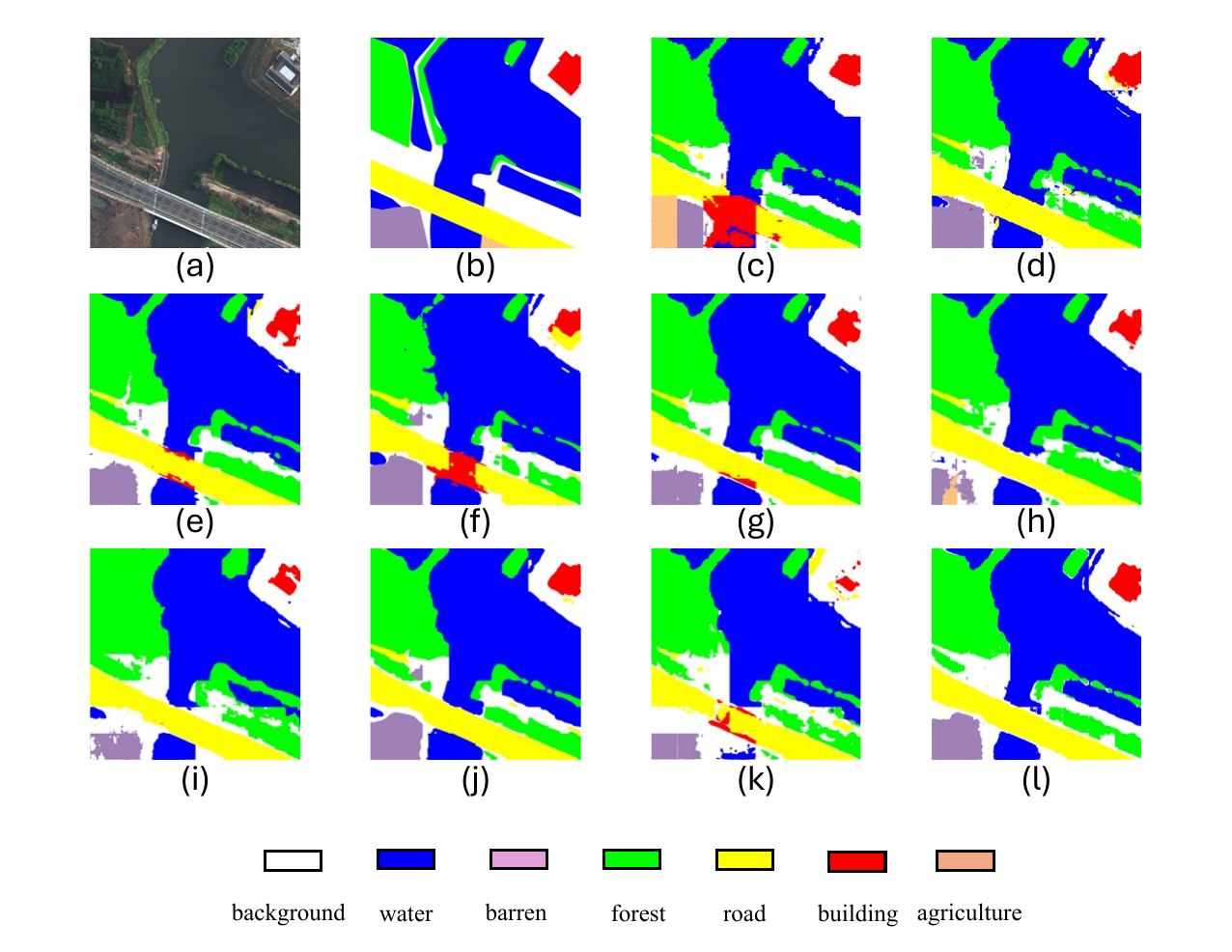}
    \caption{Performance comparisons on the LoveDA Urban dataset with the size of $1024 \times 1024$. (a) NIRRG images, (b) Ground truth, (c) ABCNet, (d) MANet, (e) CMTFNet, (f) UNetFormer, (g) FTUNetFormer, (h) HST\_UNet, (i) TransUNet, (j) RS3Mamba, (k) RS-Mamba and (l) PPMamba. }
    \label{fig6}
\end{figure*}

\begin{figure*}[t]
    \centering
    \includegraphics[width=\linewidth]{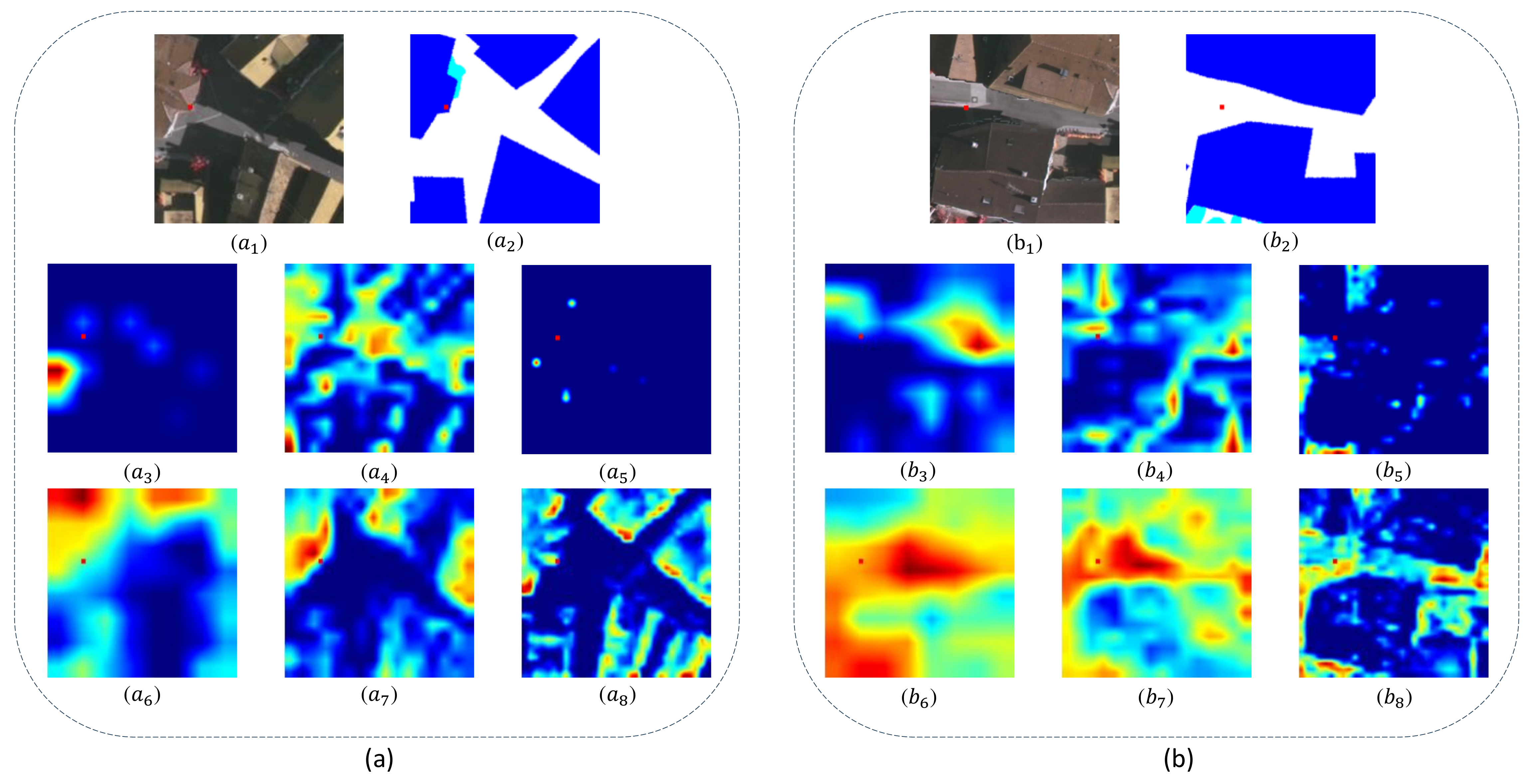}
    \caption{The comparison of heatmaps of RS-Mamba and PPMamba. ($a$)($a_1$) the NIRRG image, ($a_2$) the groud truth,  ($a_3$-$a_5$) three heatmaps from RS-Mamba, ($a_6$-$a_8$) three heatmaps from PPMamba. ($b$) are organized in the same way. ($a$) Heatmaps of encoders determine if a pixel belongs to \textit{buildings} or not. ($b$) Heatmaps of encoders determine if a pixel belongs to \textit{impervious surfaces} or not. The selected NIRRG images and ground truth are slid with a fixed window size from the ISPRS Vaihingen dataset.}
    \label{fig7}
\end{figure*}

\subsubsection{Performance comparison on ISPRS Vaihingen}
As shown in Table~\ref{tab:table1}, PPMamba demonstrated significant improvement over its baseline model, RS-Mamba. The primary evaluation metrics, mIoU and mF1, increased by $6.83\%$ and $4.63\%$, respectively, confirming that RS-Mamba has limitations in RS image semantic segmentation tasks, which PPMamba effectively overcomed. Notably, PPMamba achieved the best performance across all five foreground classes. For the impervious surface class, PPMamba achieved an F1 score of $91.86\%$, nearly $1.00\%$ higher than RS3Mamba, underscoring its ability to distinguish between urban structures and other land cover types. It also led in the building class, surpassing the baseline model by $3.42\%$. This superior performance suggests that PPMamba excels at capturing complex building shapes and boundaries, which are often challenging due to occlusions and shadows. In the low vegetation class, PPMamba outperformed ABCNet by $1.11\%$ and FTUNetFormer by $2.56\%$, highlighting its accuracy in identifying and segmenting areas covered by grass, shrubs, and other low-height vegetation. Furthermore, PPMamba achieved the highest F1 score and IoU in the tree and car categories, with an IoU of $80.03\%$, surpassing other models by at least $5\%$ and RS-Mamba by $15.79\%$. This improvement reflects its enhanced ability to recognize local features, especially when detecting cars, occupying only a small portion of the Vaihingen images. These results have demonstrated the potential of PPMamba in effectively recognizing a wide range of categories.

Fig.~\ref{fig5} presents a visual comparison of segmentation results on the ISPRS Vaihingen dataset, including outputs from all models, the NIRRG image, and the ground truth. The visual results have shown that PPMamba provided more accurate and detailed segmentation, particularly in building boundaries and tree and low vegetation regions. Notably, only PPMamba correctly identified the small building in the lower part of the image, surrounded by extensive low vegetation and trees. Additionally, PPMamba’s segmentation of buildings (blue areas) maintained continuous and precise outlines at the bottom of the image, with building boundaries seamlessly connecting to those of trees and low vegetation without any gaps. In contrast, the blue areas produced by other comparison models, including our baseline model, RS-Mamba, showed blurred and jagged edges. PPMamba also excelled in distinguishing between low vegetation and tree classes, where other models often suffered from over-segmentation.

\subsubsection{Performance comparison on LoveDA Urban}
As shown in Table~\ref{tab:table2}, experiments have been performed on the LoveDA Urban dataset as a supplementary benchmark to further validate the performance of PPMamba. Similar to the results on the previous dataset, PPMamba achieved the highest mIoU and mF1 scores among the nine state-of-the-art models. It significantly outperformed the baseline model, with an improvement of $7.90\%$ in mIoU and $7.29\%$ in mF1, due to its superior capability in capturing local features in RS images compared to the baseline model, RS-Mamba. Notably, PPMamba exhibited impressive performance in the background, building, and water categories. Specifically, PPMamba achieved an IoU of $69.27\%$ and an F1 score of $52.57\%$ in the background class, ranking among the best performers across all models. This highlights PPMamba’s ability to accurately capture background features and effectively distinguish them from adjacent categories. In the building class, our model also achieved the highest F1 ($71.25\%$) and IoU ($77.44\%$) scores, demonstrating its strength in precisely segmenting building structures. For the water category, PPMamba attained the top F1 score of $78.35\%$, outperforming the second-best model, TransUNet, by $4.67\%$, and it also led in IoU with a score of $66.87\%$. These results underscored PPMamba’s excellent ability to segment water areas. Although PPMamba achieved the second-best metrics for the road and barren categories, its performance remained highly competitive. In the road class, PPMamba’s F1 score was only $0.24\%$ lower than that of FTUNetFormer, and it is just $0.64\%$ behind RS3Mamba in the barren class. Additionally, PPMamba outperformed the baseline model, RS-Mamba, by $5.58\%$ in F1 score for the forest category, indicating notable improvement in recognizing vegetation areas.

Fig.~\ref{fig6} provides a visual comparison of the test results across all models, along with the NIRRG image and the ground truth. In the top-right corner, a square red area is clearly delineated by PPMamba, which accurately captured the square building with clear and contiguous outlines, free from significant errors. In contrast, the baseline model RS-Mamba and other state-of-the-art models such as CMTFNet and UNetFormer struggled with this task. They failed to clearly outline the square shape, with UNetFormer even misclassifying parts of the building as roads. RS-Mamba also had difficulty in detecting the yellow road area in the lower part of the image, leading to blurred boundaries between the road and building classes. This resulted in some road areas being incorrectly classified as buildings (red). In contrast, PPMamba produced continuous and precise boundaries for road areas, clearly distinguishing them from adjacent classes.

In summary, the comparative results across two different datasets have demonstrated the significant potential of PPMamba in RS image semantic segmentation, which confirms that PPMamba is more competitive and effective than both its baseline model and other state-of-the-art models mentioned in this study.

\subsection{Feature Capture Capability Comparison}
Mamba excels at capturing long-range dependencies \citep{gu2023mamba}, but its ability to extract local features is less effective. This experiment aims to analyze the differences in local feature extraction between the baseline model RS-Mamba and our enhanced model PPMamba using heatmaps. In Fig.~\ref{fig7}, the category of the red pixel at coordinates $[99, 49]$ is labeled as ``buildings" in subimage ($a$) and ``impervious surfaces" in subimage ($b$). In these heatmaps, red indicates a higher likelihood of predicting the designated category, while blue suggests little to no correlation. In the last two rows of both ($a$) and ($b$) in Fig.~\ref{fig7}, the feature maps with sizes $[1, 768, 8, 8]$, $[1, 384, 16, 16]$, and $[1, 192, 32, 32]$ are shown in the format $[B, C, H, W]$ from left to right in each row. The NIRRG images were taken from the ISPRS Vaihingen dataset, where a $256\times 256$ window was slid across the images with a set stride, generating the NIRRG images in the heatmaps.

Fig.~\ref{fig7} compares the feature extraction capabilities of RS-Mamba and PPMamba in two selected scenarios. In subimages ($a_3$)-($a_5$), RS-Mamba frequently misclassified buildings and nearby low vegetation or impervious surfaces as similar features. As a result, large patches of red and yellow were scattered across the sub image ($a_4$). In contrast, PPMamba demonstrated superior local feature extraction. In the first two sub images ($a_6$) and ($a_7$), PPMamba delineated the contours of all buildings, highlighting them with prominent red and yellow regions that closely aligned with the ground truth. Moreover, PPMamba accurately identified building outlines in subimage ($a_8$), while RS-Mamba failed to detect any building pixels. In scenario (b), RS-Mamba struggled to differentiate between impervious surfaces and buildings. In subimage ($b_4$), red and yellow regions erroneously covered the building category. On the other hand, PPMamba exhibited better performance in subimages ($b_6$)-($b_8$), accurately recognizing the shape of impervious surfaces not only in the most concrete feature map but also in the most abstract one. Table~\ref{tab:table1} further supports this analysis, showing that PPMamba achieved the highest F1 scores for both buildings ($95.94\%$) and impervious surfaces ($91.86\%$) among all state-of-the-art models. These heatmap comparisons have clearly demonstrated that PPMamba offers a more effective local feature extraction capability than its baseline model, RS-Mamba.

\subsection{Ablation Study}
To validate the effectiveness of the proposed multi-branch auxiliary architecture and pyramid-shaped convolutional blocks, six ablation experiments were conducted on both the ISPRS Vaihingen and LoveDA Urban datasets. In Table~\ref{tab:table3}, the first row for each dataset represents the baseline model RS-Mamba, which does not include the multi-branch convolutional auxiliary architecture. The second row corresponds to a version of PPMamba with four convolutional branches, but with all branches having identical kernel sizes. The final row represents the full PPMamba model, which combines the multi-branch auxiliary structure with pyramid-shaped kernel sizes for the convolutional blocks.

\begin{table}[H]
	\caption{The ablation study of PPMamba on ISPRS Vaihingen and LoveDA Urban dataset.\textbf{Bold} font represents the best values.}
	\label{tab:table3}
	\centering
	\begin{threeparttable}  
		\begin{tabular}{*6{c}}\toprule
			\textbf{Dataset} & \textbf{Model} & \textbf{MB}\tnote{1} & \textbf{PS}\tnote{2} & \textbf{mF1} & \textbf{mIoU} \\ 
			\midrule
			Vaihingen & RS-Mamba &  &  & 83.71 & 72.77 \\ 
			Vaihingen & PPMamba  & $\surd$ &  & 87.55 & 78.38 \\
			Vaihingen & PPMamba & $\surd$ & $\surd$ & \textbf{88.34} & \textbf{79.60} \\
			\midrule
			Urban & RS-Mamba &  &  & 54.47 & 38.24 \\ 
			Urban & PPMamba & $\surd$ &  & 58.38 & 42.98 \\
			Urban & PPMamba & $\surd$ & $\surd$ & \textbf{61.76} & \textbf{46.14} \\
			\bottomrule
		\end{tabular}
		\begin{tablenotes}    
			\footnotesize              
			\item[1] MB: Multi-Branch        
			\item[2] PS: Pyramid-shaped        
		\end{tablenotes}            
	\end{threeparttable}       
\end{table}

Table~\ref{tab:table3} presents the performance comparison across all three configurations. PPMamba with four identical branches showed significant improvements in evaluation metrics, increasing mIoU by $5.61\%$ for Vaihingen ($4.74\%$ for Urban), and mF1 by $3.84\%$ for Vaihingen ($3.91\%$ for Urban). These substantial enhancements indicate that introducing a multi-branch convolutional structure significantly strengthened RS-Mamba’s feature extraction capability. Furthermore, by employing varying kernel sizes of $1\times 1$, $3\times 3$, $5\times 5$, and $7\times 7$ as part of the pyramid pooling operation, PPMamba can capture local features at different scales in RS images. This resulted in further increases in mIoU and mF1 by $1.22\%$ for Vaihingen ($3.16\%$ for Urban), and $0.78\%$ for Vaihingen ($3.38\%$ for Urban), respectively. Overall, the combination of the four-branch auxiliary architecture and pyramid-shaped convolutional blocks has made PPMamba highly effective and competitive in the semantic segmentation of RS images.

\begin{table}[h]
	\begin{center}
		\caption{The computational complexity analysis. FLOPs and parameter were evaluated by a random tensor with size [1, 3, 256, 256]. Memory was evaluated by NVIDIA-SMI when running the process with batch size = 2. \textbf{Bold} font represents the best values.}
		\label{tab:table4}
		\begin{tabular}{ccccc}
			\toprule
			\textbf{Model} & \textbf{FLOPs} & \textbf{Parameter} & \textbf{Memory} & \textbf{mIoU} \\
			& \textbf{(G)} & \textbf{(M)} & \textbf{(MiB)} & \textbf{(\%)} \\
			\midrule
			ABCNet& 3.91 & 13.39 & \textbf{1068} & 74.57 \\ 
			MANet & 19.45 & 35.86 & 1744 & 76.32 \\ 
			CMTFNet& 8.57 & 30.07 & 1728 & 78.06\\ 
			UNetFormer& \textbf{2.94} & \textbf{11.68} & 1074 & 77.60 \\
			FTUNetFormer& 25.51 & 75.16 & 3156 & 75.09 \\ 
			HST\_UNet& 11.51 & 29.39 & 1926 & 78.67 \\ 
			TransUNet& 88.29 & 311.23 & 5744 & 78.78 \\ 
			RS3Mamba& 15.83 & 49.66 & 2204 & 78.04 \\ 
			RS-Mamba& 9.45 & 40.73 & 2698 & 72.77 \\ 
			\midrule
			PPMamba& 10.36 & 44.77 & 3040 & \textbf{79.60} \\ 
			\bottomrule
		\end{tabular}
	\end{center}
\end{table}

\subsection{Model Complexity Analysis}
Table~\ref{tab:table4} presents the computational complexity analysis for all the models discussed in this paper. FLOPs, parameters, and memory usage are used to comprehensively assess the complexity of PPMamba compared to other state-of-the-art models. FLOPs refers to the number of floating-point operations required to run a network model, indicating the computational load during inference. Parameters represent the number of model parameters that need to be learned, serving as an important measure of model complexity. Generally, models with more parameters have greater expressive power. Memory usage, which refers to GPU memory consumption, is influenced by both model size and batch size. In this analysis, the batch size is fixed at two, so only model size affects GPU memory usage.

From Table~\ref{tab:table4}, several insights into the complexity of PPMamba can be drawn. Firstly, PPMamba requires $10.36$ GFLOPs, making it quite competitive among the selected models. This indicates that the time complexity of PPMamba is comparable to that of some CNN-based models, owing to the fast inference speed characteristic of the Mamba architecture. This advantage allows PPMamba to outperform many transformer-based models in terms of computational efficiency. In terms of parameters, PPMamba's count is slightly higher at $44.77$ million, primarily due to the local auxiliary mechanism, which uses a four-branch pyramid-shaped structure. The pyramid-shaped convolutional blocks are designed to capture local features at multiple scales, adding to the model's complexity. While its parameter count is slightly higher than that of MANet ($35.86$ M) and CMTFNet ($30.07$ M), it remains significantly lower than models such as FTUNetFormer ($75.16$ M) and TransUNet ($311.93$ M). Given its superior mIoU performance, PPMamba is considered an excellent choice for RS image semantic segmentation tasks.

\section{Conclusion}\label{sec:conclusion}
This work has proposed a novel model called PPMamba, which integrates CNN and Mamba to address RS image semantic segmentation tasks. To mitigate the issue of local information loss, the core architecture of PPMamba, the PP-SSM block, is proposed and incorporated into the encoder. Endowed with the OSS model, the proposed PP-SSM block selectively scans feature maps in eight different directions, with a pyramid-shaped convolutional auxiliary mechanism to extract both local and global features from input images. This innovative design allows PPMamba to achieve competitive performance while maintaining linear computational complexity. To validate the effectiveness of PPMamba’s architecture, comprehensive experiments have been conducted on two widely used RS datasets, ISPRS Vaihingen and LoveDA Urban. The results have confirmed that the proposed semantic segmentation model can substantially outperform conventional models.

\small
\bibliographystyle{ieeetr}
\bibliography{references}
\end{document}